\pdfoutput=1

\documentclass[11pt]{article}

\usepackage{ACL2023}

\usepackage{times}
\usepackage{latexsym}

\usepackage[T1]{fontenc}

\usepackage[utf8]{inputenc}

\usepackage{microtype}

\usepackage{inconsolata}

\usepackage{amsmath}
\usepackage{amsfonts}
\usepackage{graphicx}
\usepackage{subcaption}
\usepackage{tikz}
\usepackage{multirow}
\usepackage{float}
\usepackage{booktabs}
\usepackage{tcolorbox}
\usepackage{float}
\usepackage{multicol}
\usepackage{algorithm}
\usepackage{algpseudocode}
\usepackage{soul}
\usepackage{colortbl}

\title{LlaMaVAE: Guiding Large Language Model Generation via Continuous Latent Sentence Spaces}

\author{Yingji Zhang$^{1\dagger}$,~ Danilo S. Carvalho$^{1}$,~ Ian Pratt-Hartmann$^{1}$,~ Andr\'{e} Freitas$^{1,2}$ \\
  Department of Computer Science, University of Manchester, United Kingdom$^{1}$ \\
  Idiap Research Institute, Switzerland$^{2}$ \\
  \texttt{\{firstname.lastname\}@[postgrad.]$^{\dagger}$manchester.ac.uk}}

\begin{document}
\maketitle
\begin{abstract}
Deep generative neural networks, such as Variational AutoEncoders (VAEs), offer an opportunity to better understand and control language models from the perspective of sentence-level latent spaces. To combine the controllability of VAE latent spaces with the state-of-the-art performance of recent large language models (LLMs), we present in this work LlaMaVAE, which combines expressive encoder and decoder models (sentenceT5 and LlaMA) with a VAE architecture, aiming to provide better text generation control to LLMs. In addition, to conditionally guide the VAE generation, we investigate a new approach based on flow-based invertible neural networks (INNs) named Invertible CVAE. Experimental results reveal that LlaMaVAE can outperform the previous state-of-the-art VAE language model, Optimus, across various tasks, including language modelling, semantic textual similarity and definition modelling. Qualitative analysis on interpolation and traversal experiments also indicates an increased degree of semantic clustering and geometric consistency, which enables better generation control.

\end{abstract}

\section{Introduction}
Large language models (LLMs) have demonstrated the ability to encode and generate text, capturing expressive sentence-level and discourse-level linguistic properties, which prompts an increasing number of studies that explore their controllability, such as via \textit{prompting} \cite{petroni-etal-2019-language, liu2021pretrain, li2021prefixtuning}. However, most \textit{prompting} approaches require carefully designed templates, and the underlying mechanisms of prompt-answer consistency are yet to be fully unveiled. 

Complementarily, latent generative models, such as Variational AutoEncoders (VAEs) \cite{kingma2013auto}, have enabled sentence-level and discourse-level latent representations for natural language, leading to better generative control in various downstream tasks, such as text style transfer \cite{john-etal-2019-disentangled} and natural language definition generation \cite{carvalho2023learning}. They also enabled advances in disentangled representation learning in natural language, where models have been demonstrated to improve the localisation of syntactic, semantic and conceptual properties within the latent space, thus allowing for improved generative control for conceptually complex sentences \cite{zhang2022, zhang2023learning,zhang2023graph,zhang2023towards}.

To leverage the strengths of both LLMs and VAEs, this paper raises the question on whether LLMs can be integrated with VAEs in order to improve the abstract-level \cite{subramanian2018towards}, sentence-level representations with better generative control. Previously, \citet{li2020optimus} explored the controllability of the latent sentence space of GPT2 \cite{Radford2019LanguageMA} proposing the Optimus architecture. In the Optimus architecture, BERT encodes natural language sentences into a continuous latent sentence space, which is then decoded via GPT2. Since the latent space is sentence-level and lower-dimensional, it can better control the generation of language models by manipulating the movement of sentence vectors over the latent space, such as traversal, interpolation \citep{bowman2016generating}, and arithmetic. However, the structure of BERT-GPT2 is gradually becoming outdated with the emergence of LLMs, such as LlaMA \cite{touvron2023llama}.

Therefore, building upon the influential research conducted by \cite{li2020optimus} and the advancements made at the interface between VAEs and LLMs, we propose a new mechanism to control LLM generation through VAE architecture. In this framework, the encoder and decoder components are comprised of sentenceT5 (sT5) \cite{https://doi.org/10.48550/arxiv.2108.08877} and LlaMA (7B) \cite{touvron2023llama}, respectively. By combining these components with a VAE latent space, we aim to leverage the strengths of both LLMs and VAEs and further enhance the capabilities of language generation and control.

We evaluate our model from three perspectives: (1) pre-training (language modelling task); (2) sentence encoding (semantic textual similarity task \cite{cer-etal-2017-semeval} and linguistic probing task \cite{conneau2018probing}); and (3) controlled decoding (guided generation via latent space geometry and definition modelling task \cite{mickus-etal-2022-semeval}). 

To adapt the VAE architecture to the definition modelling task, which aims to generate word definition given word embedding or its reversed process, we propose a novel approach to conditionally guide VAE generation via a flow-based invertible neural network (INN) \cite{dinh2014nice}, named Invertible Conditional VAE (CVAE). Since the INN mechanism has a low computational overhead and models the bijective transformation, we can flexibly transform the mapping between the word embedding and the pretrained latent sentence space from LlaMaVAE without architectural modifications and then re-train the large LlaMaVAE model. More importantly, the latent space with elastic and geometrically consistent characteristics\footnote{elastic and geometrically consistent refer to semantic behaviour of distance and vector operations in VAE latent spaces.} can weaken the information loss caused by the INN transformation, potentially resulting in better definition modelling.

Extensive experimentation shows that our model consistently surpasses the state-of-the-art LM-VAE, Optimus, on various benchmark datasets. The overview of the model architecture and the experimental setup is provided in Figure \ref{fig:setup}. Our contributions can be summarised as follows:

\textbf{1.} We integrate the VAE architecture with a large language model (Figure \ref{fig:LlaMaVAE}). In this framework, the encoder component utilises pre-trained sentenceT5, while the decoder component employs LlaMA. 

\textbf{2.} We build a pre-trained LlaMaVAE, where the hidden layers of LlaMa are frozen, on four datasets: WorldTree, WordNet, Wiktionary, and Wikipedia. This enables replication and extension of the proposed approach on distinct corpora.

\textbf{3.} We comprehensively evaluate LlaMaVAE on relevant benchmarks: Semantic Textual Similarity (STS-2012-2015, STS-B, SICK-R), Linguistic Probing \cite{conneau2018probing}, and Definition Modelling tasks (CODWOE). These evaluations consistently demonstrate performance improvements compared to Optimus. 

\textbf{4.} We propose a novel approach to conditionally guide the VAE-based generation via flow-based INN, calling it Invertible CVAE (Figure \ref{fig:definition_model}). This new mechanism expands the controllability of the VAE architecture by decoupling the decoder's dependency on the inputs of the encoder.

\section{Related work} \label{sec:related}
\paragraph{Controlling LLMs generation} Recently, effective control over LLM generation has become a priority area of research. In addition to fine-tuning LLMs via parameter-efficient adapters \cite{houlsby2019parameterefficient, he2022unified, hu2021lora}, the guided generation via \textit{prompting} optimisation \cite{liu2021pretrain} became increasingly popular. The latter entails two main varieties, including \textit{cloze prompts} \cite{petroni-etal-2019-language}, which predicts masked tokens in a textual string, and \textit{prefix prompts} \cite{li2021prefixtuning}, which continue a string prefix. Both approaches have their limitations, including the fact that \textit{cloze prompts} require carefully designed templates, while \textit{prefix prompts} have limited control of the semantic consistency of the generated text, especially in long, conceptually complex and domain-specific texts \cite{li2021prefixtuning, wysocka2023large}. In this work, we propose controlling LLM generation through the use of disentanglement mechanisms enabled by Invertible CVAEs, which has the potential to address the aforementioned issues.

\paragraph{Language VAE} In addition to Optimus \cite{li2020optimus}, most language VAE works are focused on LSTM architectures on different text generation tasks, including story generation \cite{fang2021transformerbased}, dialogue generation \cite{zhao-etal-2017-learning}, text style transfer \cite{john-etal-2019-disentangled, shen2020educating}, text paraphrasing \cite{bao-etal-2019-generating}, among other tasks. In this work, we focus on large-scale pre-trained models with the sT5-LlaMa VAE setup, and evaluate it on a definition modelling task. In addition to generation-related tasks, we also examine VAE latent sentence embeddings on sentence similarity tasks, including Semantic Textual Similarity (STS) \cite{cer-etal-2017-semeval} and linguistic probing task \cite{conneau2018probing}.

\paragraph{Invertible Neural Networks in NLP} The bijective properties of INN-based representations have recently been investigated in language. \citet{csahin2020two} concentrate on modelling morphological inflection and lemmatisation tasks, utilising an INN to learn a bijective transformation between the word surface and its morphemes. \citet{li2020sentence} focused on sentence-level representation learning, transforming sentences from a BERT sentence embedding space to standard Gaussian space, improving sentence embeddings on various semantic textual similarity tasks. Recently, \citet{zhang2023learning} explored the semantic disentanglement and separation of latent spaces with an integrated INN mechanism. This work builds upon and expands the controllability of VAE architecture by decoupling the decoder's dependency on the inputs, having the relationship between inputs, embeddings, and outputs be also invertible. 

\section{Methodology} \label{sec:latent_props}

\begin{figure}[ht!]
    \includegraphics[scale=0.35]{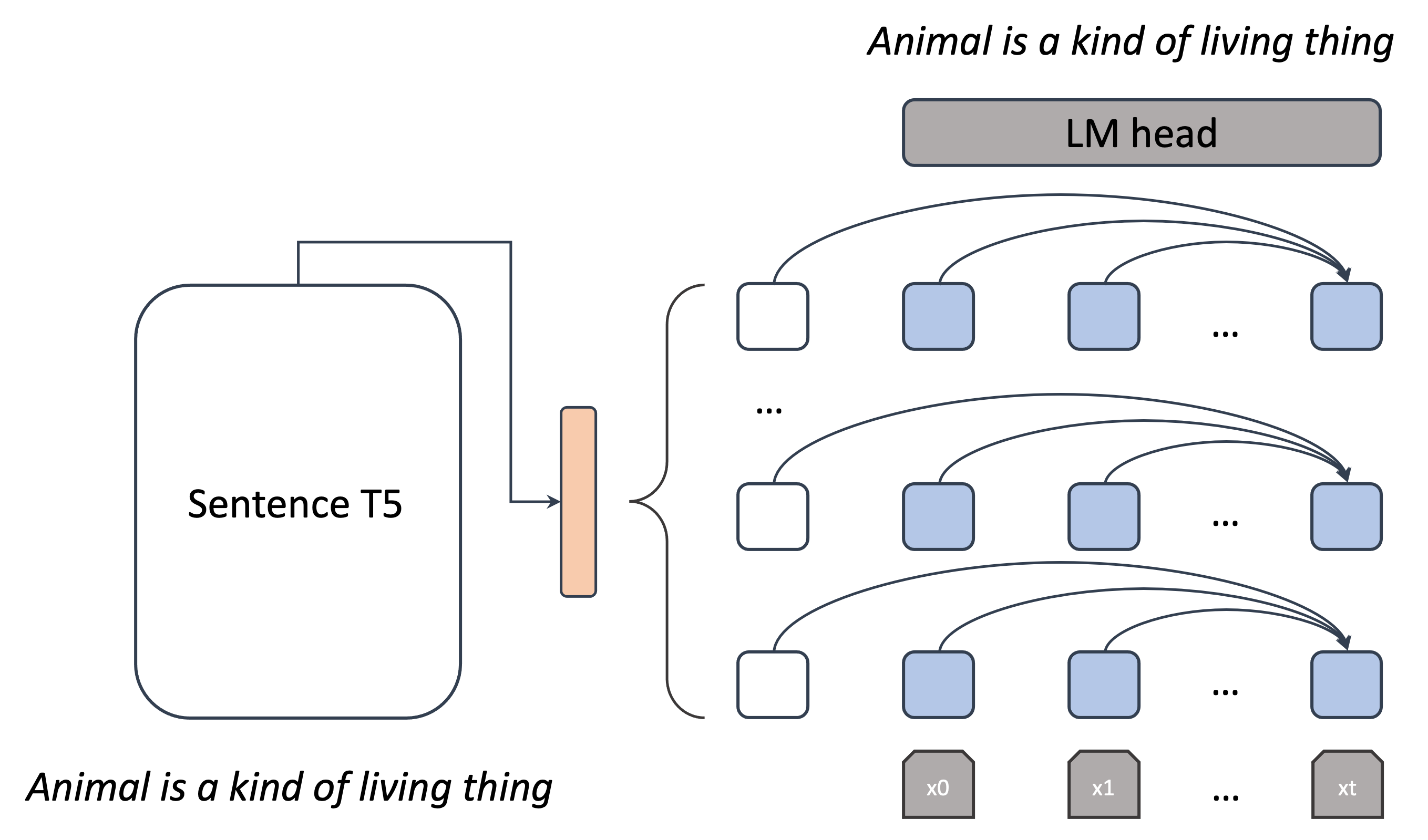}
    \caption{LlaMaVAE architecture where the hidden layers of LlaMa are frozen during pre-training.}
    \label{fig:LlaMaVAE}
\end{figure}

\paragraph{Language Modelling} When combining the VAE with sT5 and LlaMA, we adopt the "memory" setup from Optimus, in a sentence reconstruction setting. Firstly, sT5 encodes the input sentence, denoted as $x$, into the latent space $N(\mu, \Sigma)$. The parameters $\mu$ and $\Sigma$ are trainable. Next, a sample $z \sim N(\mu, \Sigma)$ is passed through a multi-layer perceptron denoted as $W$. $W$ expands the dimensionality of $z$ to obtain a fixed-length embedding $h \in R^{D \times L \times H}$. Here, $D$, $L$, and $H$ represent the dimensions of heads, the number of heads, and the number of hidden layers, respectively. In the case of LlaMa (7B), these values are 128, 32, and 32, respectively. Finally, each $v \in R^{128 \times 1 \times 1}$ is considered an additional key and value within each self-attention network of every hidden layer. This process can be summarized as follows:
\begin{gather*}
\text{MultiHead}\Big(Q, [W(z); K], [W(z); V] \Big)
\end{gather*}
where $Q$, $K$ and $V$ represent the query, key, and value hidden representations commonly used in recent LLMs. An overview of the LlaMaVAE architecture is displayed in Figure \ref{fig:LlaMaVAE}.

The LlaMaVAE can be trained via the evidence lower bound (ELBO) on the log-likelihood of the data $x$ \cite{kingma2013auto}. To avoid the KL vanishing issue, which refers to the Kullback-Leibler (KL) divergence term in the ELBO becomes very small or approaches zero, we select the cyclical schedule to increase weights of KL $\beta$ from 0 to 1 \cite{fu-etal-2019-cyclical} and a KL thresholding scheme \cite{li-etal-2019-surprisingly} that chooses the maximum between KL and threshold $\lambda$. The final objective function can be described as follows:
\begin{align*} \label{eq:elbo_loss}
\mathcal{L}_\text{VAE} = & \mathbb{E}_{q_\phi(z|x)} \Big[ \log p_{\theta} ( x | z ) \Big]  \\
& - \beta \max \left[ \lambda , \text{KL} q_\phi(z|x) || p(z) \right ]
\end{align*}

\paragraph{Definition Modelling} The definition modelling task \cite{noraset2016definition} aims to generate a definition text given a corresponding word embedding. We follow the setup of the CODWOE shared task \cite{mickus-etal-2022-semeval}, which defines two subtasks: 1. definition modelling (vector-to-definition) and 2. reversed dictionary (definition-to-vector). 

To model both transformations with one single network architecture, we select flow-based Invertible Neural Networks (INNs) \cite{dinh2014nice,dinh2016density,kingma2018glow} that define a bijective mapping between observation distribution $p(x)$ and latent distribution $p(z)$. We use $T$ and $T^{-1}$ to represent forward mapping (from $p(x)$ to $p(z)$) and backward mapping (from $p(z)$ to $p(x)$), respectively. Unlike VAEs, which approximate the posterior distribution to multivariate Gaussian distributions, INNs use multivariate Gaussian directly. The following objective function can learn the bijective mapping: 
\begin{equation}
\mathcal{L}_{INN} = 
- \mathbb{E}_{x \sim p(x)} \Big[ T(x) \Big]^2 - \log \left| T^{-1}(x) \right| \nonumber
\end{equation}
where $T(x)$ learns the transformation from $x$ to $z \sim N(0, 1)$. $\left| T^{-1}(x) \right|$ is the determinant of the Jacobian, which indicates how much the transformation locally expands or contracts the space. $- \log \left| T'(x) \right|$ ensures the integration of the probability density function is one.

The forward and reversed mapping can be performed via the \textit{coupling} layer \cite{dinh2016density, kingma2018glow}. The basic form of forward mapping can be described as follows:
$$
z = T(x) = \begin{cases} 
z_1 = x_1 \\
z_2 = x_2 ~ op ~ m_{\theta} (x_1)
\end{cases}
$$
Where $[x_1;x_2]=\text{split}(x)$, $[z_1; z_2]=\text{split}(z)$, $m_{\theta}$ is any kind of network. The reversed mapping can be obtained:
$$
x = T^{-1}(z) = \begin{cases} 
x_1 = z_1 \\
x_2 = z_2 ~ op^{-1} ~ m_{\theta} (z_1)
\end{cases}
$$
The ($op$, $op^{-1}$) are symmetrical mathematical operations in flow-based INNs, such as ($+$, $-$) and ($\odot$, $\div$). To fully utilise the representation capabilities of a pre-trained latent sentence space from the language modelling task, we share the latent spaces between INN and LlaMaVAE. That is, given a pre-trained latent sentence space (multivariate Gaussian) with parameters $\mu$ and $\Sigma$, the INN can learn the transformation between the word space and $N(\mu, \Sigma)$. The latent space with elastic and geometrically consistent characteristics can weaken the information loss caused by the INN transformation, potentially resulting in better definition modelling.

\begin{figure}[ht!]
    \includegraphics[scale=0.37]{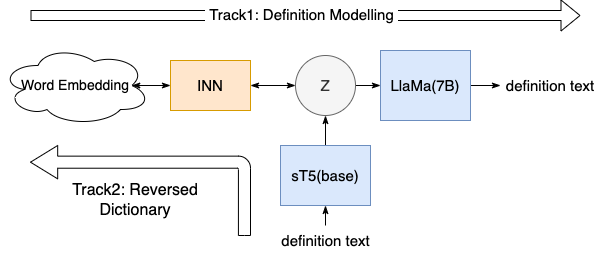}
    \caption{Definition modelling architecture.}
    \label{fig:definition_model}
\end{figure}

In this work, we train the forward and reversed transformations separately to independently evaluate both subtasks and avoid the limitations of bidirectional mapping on model performance. As for forward transformation, given (vector, definition) pair, $(w, x)$, we directly optimise the likelihood of $P(z|w) = T(w)$. $T$ represents the INN, $z \sim N(\mu, \Sigma)$, $\mu$ and $\Sigma$ can be obtained via $E(x)$ where $E$ is the Encoder. The objective function can be described as follows: 
\begin{equation}
\begin{split}
\mathcal{L}_{forward} = 
& - \mathbb{E}_{(x, w) \sim p(x, w)} \frac{\Big[ T(w) - E_{\mu}(x) \Big]^2}{E_{\Sigma}(x)}\\ \nonumber
\end{split}
\end{equation}
As for reversed transformation, we optimize the inverted INN, $T^{-1}$, via mean square error (MSE):
\begin{equation}
\begin{split}
\mathcal{L}_{reverse} = 
& - \mathbb{E}_{(x, w) \sim p(x, w)} \Big[ T^{-1}(E(x)) - w \Big]^2\\ \nonumber
\end{split}
\end{equation}
Figure \ref{fig:definition_model} visualises the process of INN and LlaMaVAE in the definition modelling task.

\begin{figure}[ht!]
\centering
    \includegraphics[scale=0.3]{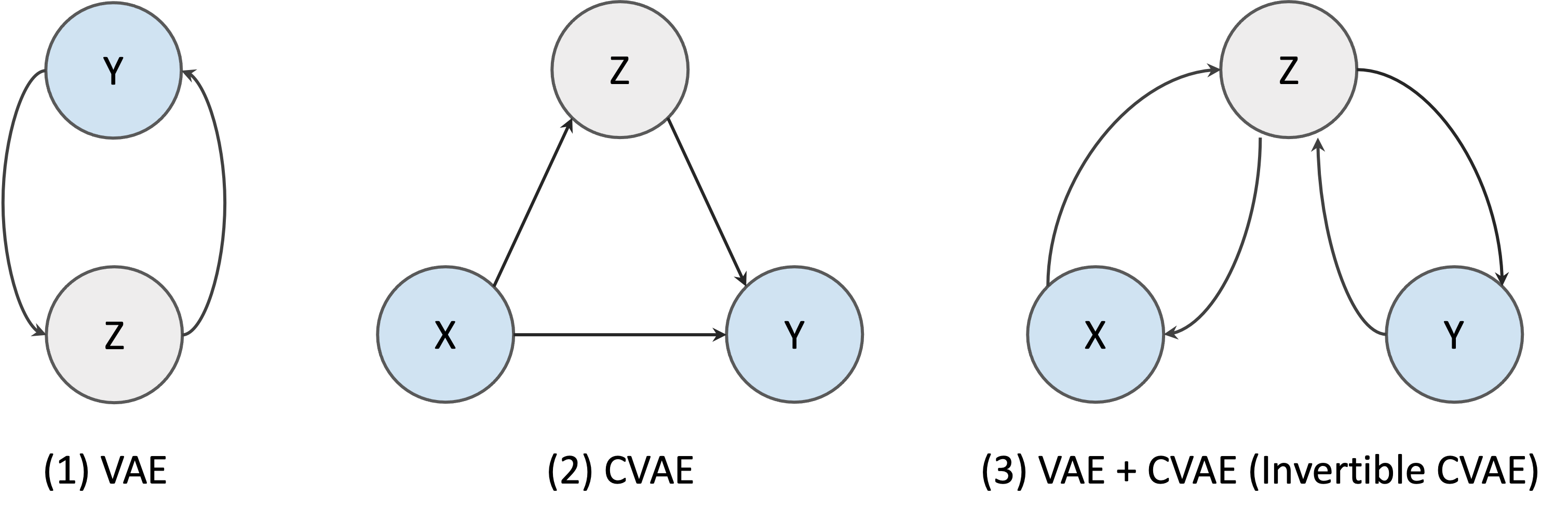}
    \caption{Computational graph where X, Y, and Z are input, output, and latent spaces.}
    \label{fig:computate_graph}
\end{figure}

\paragraph{Invertible CVAE} More generally, our approach integrates the INN mechanism to CVAEs \cite{zhao-etal-2017-learning}, naming it Invertible CVAE. Figure \ref{fig:computate_graph} illustrates the computational graphs of VAE, CVAE, and Invertible CVAE (ours). Compared with VAE $P(z) \rightarrow P(y)$, the Invertible CVAE $P(y, z|x)$ can deliver better semantic control conditioned on $x$. When compared with CVAE $P(x, z) \rightarrow p(y)$, it can maintain the ability of language modelling, which can be pre-trained at a large scale in an unsupervised manner.
\paragraph{Training setup} The latent dimension used in all experiments for VAE models is 768. Regarding the LlaMaVAE, we use the pre-trained weights of sT5(base, mean\footnote{mean: the sentence embedding is defined as the average of the encoder outputs across all input tokens}) and LlaMA(7B) as the initial weights. During training, we only fine-tune the embedding and language model head layers of LlaMA for the first epoch to inject two additional special tokens, namely <BOS> and <EOS>, and shape the output on the target corpus. The encoder and sentence space are trained throughout all epochs. The value of $\lambda$ is set to 1 for both Optimus and LlaMaVAE. Information about Optimus and the INN implementation is provided in Appendix \ref{sec:train_details} and \ref{sec:inn}. 
\section{Experiments} \label{sec:empirical}

\subsection{Language Modelling}
\paragraph{Pre-training data} We conduct a pre-training process on four corpora with different sizes. Table \ref{tab:stats_data} describes the datasets in details.
\begin{table}[ht!]
    \small
    \resizebox{7.7cm}{!}{
    \centering
    \renewcommand\arraystretch{1.3}
    \begin{tabular}{|l|ccc|}
        \hline
        Dataset & Num sents. & Avg. length & Version \\ \hline
        WorldTree & 11,430 & 9 &  \cite{jansen2018worldtree} \\
        Wordnet & 93,699 & 9 & WordNet 3.0 \\
        Wiktionary & 464,243 & 8 & Dec, 2016 \\
        Wikipedia & 1,500,323 & 12 & Dec, 2016 \\
        \hline
    \end{tabular}
    }
    \caption{Statistical information of pertaining corpora.} \label{tab:stats_data}
\end{table}
\begin{table*}[ht!]
\setlength\tabcolsep{2.5pt}
\resizebox{16.5cm}{!}{
\small
\centering
\renewcommand\arraystretch{1.5}
\begin{tabular}{c|c|rrrr|rrrr|rrrr|rrrr}
\toprule
\multirow{2}{*}{Baseline} & \multirow{2}{*}{beta} & \multicolumn{4}{c|}{WorldTree} & \multicolumn{4}{c|}{WordNet} & \multicolumn{4}{c|}{Wikipedia} & \multicolumn{4}{c}{Wiktionary}\\
 & & BLEU & BLEURT & Cosine & Loss $\downarrow$ & BLEU & BLEURT & Cosine & Loss $\downarrow$ & BLEU & BLEURT & Cosine & Loss $\downarrow$ & BLEU & BLEURT & Cosine & Loss $\downarrow$ \\ \hline
 \multirow{4}{*}{\shortstack{Optimus \\ (BERT-GPT2)}} & 0.0 & 0.21 & -0.01 & 0.78 & 1.67 & 0.67 & 0.44 & 0.96 & 0.47 & 0.65 & 0.27 & 0.97 & 0.46 & 0.63 & 0.53 & 0.97 & 0.44\\
  & 0.1 & 0.38 & -0.34 & 0.87 & 1.41 & 0.56 & 0.05 & 0.93 & 1.16 & 0.56 & 0.06 & 0.95 & 0.92 & 0.51 & 0.01 & 0.93 & 1.07 \\
  & 0.5 & 0.36 & -0.47 & 0.85 & 1.50 & 0.52 & -0.02 & 0.93 & 1.38 & 0.54 & 0.06 & 0.94 & 1.07 & 0.49 & 0.04 & 0.93 & 1.22 \\
  & 1.0 & 0.10 & -1.24 & 0.75 & 2.03 & 0.45 & -0.28 & 0.91 & 1.73 & 0.54 & 0.04 & 0.94 & 1.09 & 0.48 & -0.06 & 0.93 & 1.39 \\ \hline
  \multirow{4}{*}{\shortstack{LlaMaVAE \\ (sT5-LlaMa)}} & 0.0 & \textbf{\textcolor{blue}{0.58}} & \textbf{\textcolor{blue}{-0.01}} & \textbf{\textcolor{blue}{0.91}} & \textbf{\textcolor{blue}{0.63}} & \textbf{\textcolor{blue}{0.83}} & \textbf{\textcolor{blue}{0.69}} & \textbf{\textcolor{blue}{0.97}} & \textbf{\textcolor{blue}{0.38}} & \textbf{\textcolor{blue}{0.83}} & \textbf{\textcolor{blue}{0.60}} & \textbf{\textcolor{blue}{0.97}} & \textbf{\textcolor{blue}{0.36}} & \textbf{\textcolor{blue}{0.79}} & \textbf{\textcolor{blue}{0.55}} & \textbf{\textcolor{blue}{0.97}} & \textbf{\textcolor{blue}{0.41}} \\
  & 0.1 & 0.56 & -0.06 & 0.90 & 0.66 & 0.68& 0.22& 0.93 & 0.52 & 0.77 & 0.37 & 0.94 & 0.42 & 0.64 & 0.01 & 0.90 & 0.58 \\
  & 0.5 & 0.55 & -0.07 & 0.90 & 0.67 & 0.67 & 0.18 & 0.93 & 0.53 & 0.79 & 0.38 & 0.94 & 0.43 & 0.62 & 0.01 & 0.90 & 0.59 \\
  & 1.0 & 0.53 & -0.10 & 0.90 & 0.67 & 0.66 & 0.17 & 0.92 & 0.54 & 0.75 & 0.32 & 0.94 & 0.43 & 0.60 & -0.04 & 0.89 & 0.60\\ \hline \hline

  AAE & - & \textbf{0.35} & \textbf{-0.95} & \textbf{0.80} & \textbf{3.35} & \textbf{0.53} & \textbf{-0.57} & \textbf{0.87} & \textbf{2.31} & \textbf{0.65} & \textbf{-0.12} & \textbf{0.96} & \textbf{1.07} & \textbf{0.53} & \textbf{-0.75} & \textbf{0.84} & \textbf{1.98} \\
  LAAE & - & 0.26& -1.07& 0.78& 3.71&0.26& -1.05& 0.78& 2.62& 0.49 & -0.43&0.87&1.72& 0.40 & -0.95 & 0.81 &2.56 \\
  DAAE & - & 0.22& -1.26& 0.76& 4.00&0.17& -1.17& 0.76&2.97 &0.54 & -0.35& 0.89& 1.57& 0.42 & -0.96 & 0.80 & 2.46 \\
  $\beta$-VAE & 0.5 & 0.06& -1.14& 0.77& 3.69& 0.04 & -0.98 & 0.75 & 3.12 & 0.18& -0.96& 0.75& 2.30& 0.19 & -1.13 & 0.77 & 3.28 \\ \toprule

\end{tabular}
}
\caption{Pre-training evaluation on test set. AAE: adversarial autoencoder, LAAE: label adversarial autoencoder, DAAE: denoising adversarial autoencoder. The highest score of large VAE models and other baselines are highlighted in blue and in bold separately. Same for the remaining tables.} \label{tab:language_model}
\end{table*}
\begin{table*}[ht!]
\setlength\tabcolsep{2.5pt}
\small
\centering
\renewcommand\arraystretch{1}
\begin{tabular}{lccccccc}
\toprule
\multicolumn{2}{l}{Dataset} & STS-B & SICK-R & STS-12 & STS-13 & STS-14 & STS-15 \\ \hline \hline
\multicolumn{8}{c}{\textit{published in \cite{ethayarajh2019contextual}}} \\
\multicolumn{2}{l}{AVG. GolVe embeddings} & 58.02 & 53.76 & 55.14 & 70.66 & 59.73 & 68.25 \\ 
\multicolumn{2}{l}{AVG. BERT embeddings} & 46.35 & 58.40 & 38.78 & 57.98 & 57.98 & 63.15 \\ 
\multicolumn{2}{l}{BERT CLS embeddings} & 16.50 & 42.63 & 20.16 & 30.01 & 20.09 & 36.88 \\ \hline 

\multicolumn{8}{c}{\textit{published in \cite{li2020sentence}}} \\
\multicolumn{2}{l}{BERT(base)} & 47.29 & 58.21 & 49.07 & 55.92 & 54.75 & 62.75 \\ 
\multicolumn{2}{l}{BERT(large)} & 46.99 & 53.74 & 46.89 & 53.32 & 49.27 & 56.54 \\ 
\multicolumn{2}{l}{BERT(base)-flow} & 70.72 & \textbf{63.11} & 63.48 & 72.14 & 68.42 & 73.77 \\ 
\multicolumn{2}{l}{BERT(large)-flow} & \textbf{72.26} & 62.50 & \textbf{65.20} & \textbf{73.39} & \textbf{69.42} & \textbf{74.92} \\ \hline \hline
\multicolumn{8}{c}{\textit{Our implementation}} \\
\multirow{2}{*}{Optimus (BERT-GPT2)}& 0.0 & 50.61 & 61.88 & 25.57 & 26.92 & 33.79 & 39.42 \\
& 1.0 & 15.48 & 30.19 & 23.32 & 16.56 & 23.14 & 30.87 \\
\multirow{2}{*}{LlaMaVAE (sT5-LlaMa)}& 0.0 & \textcolor{blue}{\textbf{62.50}} & \textcolor{blue}{\textbf{69.62}} & \textcolor{blue}{\textbf{45.31}} & \textcolor{blue}{\textbf{41.73}} & \textcolor{blue}{\textbf{49.44}} & \textcolor{blue}{\textbf{57.73}}\\
& 1.0 & 28.10 & 33.94 & 30.44 & 27.25 & 37.08 & 42.95 \\ \toprule
\end{tabular}
\caption{Spearman’s correlation coefficients ($\times 100$) to evaluate Semantic textual similarity (STS).} \label{tab:semantic_similarity_autoencoding}
\end{table*}
\paragraph{Baselines} In our implementation, we utilise LlaMaVAE and Optimus \cite{li2020optimus} and incorporate four LSTM-based language autoencoding models: $\beta$-VAE \cite{Higgins2016betaVAELB}, adversarial AE (\citet{makhzani2016adversarial}, AAE), label adversarial AE (\citet{rubenstein2018latent}, LAAE), and denoising adversarial autoencoder (\citet{shen2020educating}, DAAE). Each of them has a latent sentence embedding size of 768. All supporting code and deployment documentation for the experimental pipeline will be available for reproducibility purposes in an anonymised link.

\paragraph{Quantitative evaluation} We quantitatively evaluate reconstruction on the test set via four metrics, including BLEU \cite{Papineni02bleu:a}, BLEURT \cite{https://doi.org/10.48550/arxiv.2004.04696}, cosine similarity from pre-trained sT5 \cite{https://doi.org/10.48550/arxiv.2108.08877}, and cross-entropy (Loss). The results are presented in Table \ref{tab:language_model} where the highest scores of large VAE models and LSTM-based VAE models are separately highlighted in blue and in bold, and it can be observed that (1) LlaMaVAE achieves better performance compared to other baseline models across all datasets, (2) The performance of the models decreases as the $\beta$ increases, where 0.5 shows a good trade-off point.
\begin{table*}[ht!]
\setlength\tabcolsep{2.5pt}
\small
\centering
\renewcommand\arraystretch{1}
\begin{tabular}{lccccccccccc}
\toprule
\multicolumn{2}{c}{Properties} & SentLen & WordContent & TreeDepth & TopConst & BShift & Tense & SubjNum & ObjNum & SOMO & CoordInv\\ \hline
\multicolumn{12}{c}{\textit{publised in \cite{conneau-etal-2018-cram}}} \\
Bi-LSTM AE & & \textbf{99.3} & 23.3 & 35.6 & \textbf{78.2} & \textbf{62.0} & 84.3 & \textbf{84.7} & \textbf{84.1} & 49.9 & \textbf{65.1} \\
BoV-FastText & & 66.6 & \textbf{91.6} & \textbf{37.1} & 68.1 & 50.8 & \textbf{89.1} & 82.1 & 79.8 & \textbf{54.2} & 54.8 \\ \hline \hline
\multicolumn{12}{c}{\textit{our implementation}} \\
\multirow{2}{*}{\shortstack{Optimus \\ (BERT-GPT2)}}& 0.0 &  55.8 & \textcolor{blue}{\textbf{31.2}} & 25.3 & 43.3 & \textcolor{blue}{\textbf{66.0}} & 75.8 & 73.0 & \textcolor{blue}{\textbf{76.8}} & 49.3 & \textcolor{blue}{\textbf{55.7}} \\
& 1.0 & 33.8 & 7.0 & 20.8 & 22.1 & 56.5 & 66.4 & 61.9 & 65.9 & 50.0 & 52.2\\
\multirow{2}{*}{\shortstack{LlaMaVAE \\ (sT5-LlaMa)}}& 0.0 & \textcolor{blue}{\textbf{75.2}} & 24.1 & \textcolor{blue}{\textbf{27.9}} & \textcolor{blue}{\textbf{57.2}} & 56.6 & \textcolor{blue}{\textbf{77.4}} & \textcolor{blue}{\textbf{77.8}} & 74.8 & \textcolor{blue}{\textbf{50.9}} & 53.2 \\
& 1.0 & 60.6 & 6.3 & 22.1 & 35.5 & 53.2 & 68.2 & 65.8 & 62.4 & 50.5 & 51.2 \\ \toprule
\end{tabular}
\caption{Accuracies: probing linguistic properties of latent sentence space \cite{conneau-etal-2018-cram} (AE: autoencoder).} \label{tab:linguistic_prob}
\end{table*}

\subsection{Latent Sentence Space}
\paragraph{Semantic textual similarity} Following the pre-training stage, we select Optimus and LlaMaVAE models pre-trained on the Wikipedia corpus and evaluate both models' performance on semantic textual similarity (STS) tasks across 6 datasets without fine-tuning. Those datasets include the STS benchmark (STS-B) \cite{cer-etal-2017-semeval}, the SICK-Relatedness (SICK-R) dataset \cite{marelli-etal-2014-sick}, and the STS tasks 2012-2015 \cite{agirre-etal-2012-semeval, agirre-etal-2013-sem, agirre-etal-2014-semeval, agirre-etal-2015-semeval}. All datasets were obtained via the SentEval toolkit \cite{conneau2018senteval}. We compare both models with several baselines on sentence retrieval tasks, including GloVe, BERT, and BERT-flow \cite{li-etal-2020-sentence}. The evaluation is conducted by quantitatively measuring Spearman's correlation coefficients between the predicted cosine similarity scores and the gold similarity scores provided by the datasets.

Based on the information provided in Table \ref{tab:semantic_similarity_autoencoding}, it is evident that LlaMaVAE outperforms Optimus across all datasets and achieves the highest scores on the SICK-R dataset. However, it is worth noting that both VAE-based models do not perform comparably to BERT-flow \cite{li-etal-2020-sentence} on the other datasets. Therefore, we next probe what linguistic information is lost in the latent sentence embedding of the VAE setup.

\paragraph{Linguistic properties} \citet{conneau2018probing} put forward 10 probing tasks and corresponding datasets designed to capture linguistic features of sentence representations. For each task, a classifier is trained over the corresponding dataset, where its input is the sentence embedding. Its accuracy on the test set implies the importance of the related language properties to the autoencoding task. We refer to \cite{conneau-etal-2018-cram} for an in-depth description of those properties.

As illustrated in Table \ref{tab:linguistic_prob}, we can observe that (1) LlaMaVAE can outperform Optimus on 6 out of 10 tasks, and (2) autoencoding models cannot perform well on the WordContent task, indicating that the sentence space of the autoencoder architecture does not contain the word content information, as such information has been delegated to the decoder. In LlaMaVAE, for example, the decoder uses a Byte-Pair Encoding (BPE) model, which defines its own vocabulary space. The similarity experiment shows that the semantic information contained within the latent sentence space is insufficient to deliver a granular, content-based encoding, which is closely related to the model structure and training objectives. This established a clear role for the VAE bottleneck component, which is to facilitate the syntactic and semantic coherence control (e.g. the relationship between predicate, associated arguments and their topic coherence) of the generated sentences, which is the focus of the next section.



\subsection{Natural Language Generation}
\paragraph{Guided generation via geometrical properties} Since the VAE architectures learn the mapping between sentence-level and word-level spaces, we can manipulate the movement of sentence representations to control word-level generation. Firstly, we evaluate the target VAE models via sentence interpolation, which can be described as $ z_t = z_1 \cdot (1 - t) + z_2 \cdot t  $  with $ t $ increased from $ 0 $ to $ 1 $ by a step size of $ 0.1 $ where $ z_1 $ and $ z_2 $ represent latent vectors of source and target sentences, respectively. If the latent space has consistent geometric properties and high continuity, intermediate sentences should change in content according to the semantic variation between the source and target and the size of the steps taken. An example of LlaMaVAE is illustrated in Table \ref{tab:interpolation}. The interpolation of Optimus can be found in Appendix \ref{sec:interpolate_more} (Table \ref{tab:interpolation_optimus}). Additional representative interpolation outputs are provided in Appendix \ref{sec:interpolate_more}. 
\begin{table}[ht!]
\centering
\begin{tcolorbox}[fontupper=\small, fontlower=\small]
\textcolor{blue}{Source: Mars contains ice} \\
0: Mars contains ice \\
1: Mars is a planet \\
2: Mars is made of rock \\
3: Mars is a kind of object \\
4: mars is a kind of object \\
5: oxygen is a kind of substance \\
6: oxygen is a kind of substance \\
7: milk is a kind of substance \\
8: food is a kind of substance \\
9: food is a kind of substance \\
\textcolor{blue}{Target: food is a kind of substance}
\end{tcolorbox}
\caption{LlaMaVAE: latent interpolation where IS metric is 0.30. Optimus outputs are provided in Table \ref{tab:interpolation_optimus}.}
\label{tab:interpolation}
\end{table}

Moreover, we quantitatively evaluate the smoothness of the interpolation path via the interpolation smoothness (IS) metric. It first calculates the aligned semantic distance between the source and the target (ideal semantic distance). Next, the sum of the aligned semantic distances between each pair of adjacent sentences along the path is calculated (actual semantic distance). Finally, the smoothness is determined by dividing the ideal semantic distance by the actual semantic distance. If the result is 1, the actual path is the same as the ideal, indicating more consistent geometric properties. Otherwise, this path might be more tortuous, indicating inconsistent geometric properties. The aligned semantic distance is calculated via Word Mover’s Distance \cite{zhao-etal-2019-moverscore}. Additional details on the IS metric are provided in Appendix \ref{sec:interpolate_more}. Table \ref{tab:interpolation_smoothness} illustrates the interpolation smoothness. We can observe that the LlaMaVAE has the potential to perform better latent space geometry than Optimus.
\begin{table}[ht!]
\scriptsize
\setlength\tabcolsep{2.5pt}
\small
\centering
\resizebox{7.8cm}{!}{
\renewcommand\arraystretch{1}
\begin{tabular}{c|c|cccc}
\toprule
Baseline & beta & WorldTree & Wordnet & Wikipedia & Wiktionary \\ \hline
\multirow{2}{*}{Optimus} & 0.0 & 0.16 & 0.18 & 0.22 & \textcolor{blue}{\textbf{0.17}} \\
& 1.0 & 0.13 & 0.15 & 0.21 & 0.17 \\
\multirow{2}{*}{LlaMaVAE} & 0.0 & \textcolor{blue}{\textbf{0.20}} & \textcolor{blue}{\textbf{0.20}} & \textcolor{blue}{\textbf{0.23}} & 0.15 \\ 
& 1.0 & 0.20 & 0.17 & 0.22 & 0.12 \\ \toprule
\end{tabular}
}
\caption{IS: ideal path $\div$ actual path.} \label{tab:interpolation_smoothness}
\end{table}

Furthermore, we qualitatively evaluate the geometric properties of LlaMaVAE latent space via traversal. In its latent space, where each dimension follows a Gaussian distribution, the traversal can be done by decoding the latent vector in which each dimension is resampled. Given an input, we can traverse the neighbouring points within a circle boundary where the radius is the hyperparameter. In the experiment, the radius is the \textit{L2} norm distance (500) around the input. As displayed in Table \ref{tab:traversal}, we can observe that the traversed results around a given input have similar content and are potentially factual since we froze the hidden layers of LlaMA during training, which indicates the latent sentence space has the potential to be applied in other downstream tasks, such as fact probing (a.k.a. fact retrieval) which aims to probe how much factual knowledge the LLM’s internal representations bear \cite{jiang-etal-2020-x}, and to generate controlled fact augmentation to assist Natural Language Inference models \cite{dhingra2020differentiable, valentino2021hybrid}.
\begin{table}[ht!]
\centering
\begin{tcolorbox}[fontupper=\small, fontlower=\small]
\textcolor{blue}{Input: animal is a kind of living thing} \\

1: an animal is a kind of organism \\
2: a bird is a kind of living thing \\
3: sea turtle is a kind of animal \\
4: a human is a kind of animal \\
5: sensing is a kind of animal characteristic \\
6: a butterfly is a kind of living thing \\
7: fish is a kind of living thing \\
8: frog is a kind of animal \\
9: living things are a kind of organism \\
10: a seaweed is a kind of plant
\end{tcolorbox}
\caption{LlaMaVAE: latent traversal.}
\label{tab:traversal}
\end{table}
\begin{table*}[ht!]
\setlength\tabcolsep{2.5pt}
\small
\centering
\renewcommand\arraystretch{1}
\begin{tabular}{c|c|cccccc}
\toprule
\multirow{2}{*}{Model} & \multirow{2}{*}{WordEmbed} &  \multicolumn{3}{c}{Track1: Definition Modelling} & \multicolumn{3}{c}{Track2: Reversed Dictionary}\\
&& INN loss$\downarrow$ & Sense-BLEU & MoverScore & MSE (INN loss)$\downarrow$ & Cosine & Ranking$\downarrow$ \\\hline
\multicolumn{8}{c}{\textit{Publised in \cite{mickus-etal-2022-semeval}}} \\
\multirow{3}{*}{baselines} & Electra & - & \textbf{0.0315} & \textbf{0.0673} & 1.4128 & \textbf{0.8428} & 0.4989 \\
& Char & - & 0.0263 & 0.0453 & \textbf{0.1477} & 0.7900 & 0.5021 \\
& SGNS & - & 0.0304 & 0.0830 & 0.9109 & 0.1513 & \textbf{0.4903} \\ \hline \hline
\multicolumn{8}{c}{\textit{Evaluating Invertible CVAE framework}} \\
\multirow{3}{*}{\shortstack{LlaMaVAE \\ Flow(tr)}} & Electra & \textbf{\textcolor{blue}{165.7715}} & \textbf{\textcolor{blue}{0.0269}} & \textbf{\textcolor{blue}{0.5430}} & 1.2024 & \textbf{\textcolor{blue}{0.8464}} & 0.4355  \\
& Char & 178.6500 &0.0249 & 0.5349 & \textbf{\textcolor{blue}{0.1376}} & 0.8046 & 0.4369 \\
& SGNS & 171.0692 &0.0255 & 0.5425 & 0.9467 & 0.3010 & \textbf{\textcolor{blue}{0.2235}} \\
\multirow{3}{*}{\shortstack{Optimus \\ Flow(tr)}} & Electra & 242.6433 & 0.0089 & 0.5042 & 3.4214 & 0.0090 & 0.4883 \\
& Char & 258.6515 & 0.0173 & 0.5185 & 0.4661 & 0.0062 & 0.5140 \\
& SGNS & 249.5961 & 0.0150 & 0.5161 & 1.1690 & 0.0009 & 0.5001 \\ \toprule
\end{tabular}
\caption{CODWOE shared task. More details about metrics and baselines can be found in \cite{mickus-etal-2022-semeval}.} \label{tab:definition_model}
\end{table*}
\begin{table*}[ht!]
\setlength\tabcolsep{2.5pt}
\small
\centering
\renewcommand\arraystretch{1}
\begin{tabular}{c|c|cccccc}
\toprule
\multirow{2}{*}{Model} & \multirow{2}{*}{WordEmbed} &  \multicolumn{3}{c}{Track1: Definition Modelling} & \multicolumn{3}{c}{Track2: Reversed Dictionary}\\
&& INN loss$\downarrow$ & Sense-BLEU & MoverScore & MSE (INN loss)$\downarrow$ & Cosine & Ranking$\downarrow$ \\\hline
\multirow{3}{*}{\shortstack{LlaMaVAE \\ Flow(all)}} & Electra & \textbf{\textcolor{blue}{006.0003}} & \textbf{\textcolor{blue}{0.1935}} & \textbf{\textcolor{blue}{0.6679}} & 0.3206 & 0.9340 & 0.0748 \\
& Char & 061.2831 & 0.1099 & 0.6095 & \textbf{\textcolor{blue}{0.0390}} & \textbf{\textcolor{blue}{0.9474}} & \textbf{\textcolor{blue}{0.0137}} \\
& SGNS & 061.0591 & 0.1136 & 0.6109 & 0.2296 & 0.5876 & 0.0653 \\
\multirow{3}{*}{\shortstack{Optimus \\ Flow(all)}} & Electra & 068.8520 & 0.0115 & 0.5083 & 3.4251 & 0.0039 & 0.4852 \\
& Char & 211.9257 & 0.0161 & 0.5188 & 0.4664 & 0.0003 & 0.5129 \\
& SGNS & 181.4583 & 0.0174 & 0.5208 & 1.1687 & 0.0011 & 0.4988 \\ \toprule
\end{tabular}
\caption{Target performance of Invertible CVAE framework.} \label{tab:definition_model_1}
\end{table*}

\paragraph{Definition Modelling} Finally, we analyse the performance of the model on the definition modelling task, as introduced by \cite{mickus-etal-2022-semeval} (CODWOE dataset). The aim is to evaluate the performance of the model over `conceptually dense' sentences, evaluating the model in the direction of controlled conceptual abstractions. Definitions cover a significantly different linguistic space when compared to sentence similarity tasks, which tend to concentrate on more concrete, event description-type sentences. CODWOE provides three different word embedding collections: word2vec model \cite{mikolov2013efficient} (SGNS), the ELECTRA model of \cite{clark-etal-2020-pre} (Electra), and character-based embeddings (Char). All of them have 256 dimensions. 

The model was extended with an Invertible CVAE setting in order to bridge the VAE-based models to the assigned word-embedding spaces. Since the INN needs the same input and output dimensions, to make the provided word embedding and pre-trained sentence embedding (768) have the same dimension, we repeat their word embedding three times and consider the concatenated embedding as the input of INN at the training stage. At the evaluation stage, we split the predicted embedding and calculate the mean as the result.

As for evaluation metrics on the forward definition modelling (vector-to-definition) subtask, in addition to the official metrics, including BLEU (sense-BLEU) and MoverScore \cite{zhao-etal-2019-moverscore}, we also evaluate the INN performance according to its default loss. For reversed definition modelling (definition-to-vector), MSE, Cosine Similarity, and Ranking \cite{mickus-etal-2022-semeval} are reported, where Ranking is the ratio of the number of cosine values between predicted embedding and all other golden embeddings greater than the cosine value of predicted and golden embeddings.

We fine-tune both LlaMaVAE and Optimus from the Wikipedia corpus on the training dataset. We evaluate the performance of INN models with the only training set in Table \ref{tab:definition_model}. We denote this configuration as Flow(tr). We also provide the performance of INN models learned over the full target dataset, called Flow(all), in Table \ref{tab:definition_model_1} to check whether the latent space contains enough semantic information for the definition modelling task. We examine the Invertible CVAE setup for both Optimus and LlaMaVAE where $\beta$ is 1.0 as the INN originally aims to learn the mapping between an unknown distribution and standard Gaussian distribution and provide a comparison with baselines \cite{mickus-etal-2022-semeval}.

As illustrated in Table \ref{tab:definition_model} and \ref{tab:definition_model_1}, we can observe that (1) LlaMaVAE can outperform Optimus on both tracks, (2) both Optimus and LlaMaVAE can significantly outperform baselines on the official MoverScore metric, (3) LlaMaVAE with Flow(tr) outperform baselines on Reversed Dictionary track, and on the MoverScore metric of track 1, (4) Flow(tr) substantially underperforms its target Flow(all), indicating a limitation on the generalisation capacity of the current INN architecture, despite evident decoder capacity. Those results indicate that the Invertible CVAE framework can assist the definition modelling task. The transformation between word embedding and definition text can be elastically transformed with the help of latent spaces where adjacent points have similar semantics. This elasticity can alleviate the information loss from the INN.

\section{Conclusions} \label{sec:concl}   
In this work, we present a new mechanism to control LLM generation and a large pre-trained language VAE obtained through such approach: LlaMaVAE, where the encoder is a pre-trained sentenceT5(base), and the decoder is the pre-trained LlaMA(7B), targeting for providing semantic control to recent LLMs by manipulating the sentence-level latent spaces. To keep the pre-trained knowledge and shape the output of LlaMA on new datasets, we fix the hidden layers of the decoder and only fine-tune the embedding and encoder model head layers. In addition, we extend the architecture with a new flow-based INN layer (Invertible CVAE) to support the alignment with word embeddings. 

We evaluate our model from three stages: pre-training, encoding, and decoding. In each stage, we select datasets with distinct semantic properties and compare LlaMaVAE with the state-of-the-art reference (Optimus). Experimental results indicate that our model can learn better sentence embeddings from a text generation control point of view. In the future, we will investigate the utilisation of LlaMaVAE in other downstream tasks.


\section{Limitations} \label{sec:limit}
\begin{enumerate}
    \item The exploration of different LLMs like LlaMA(65B) and GPT3 has been limited due to computational resource constraints. Whether bigger LLMs can further improve the VAE performance should be explored in the future.
    
    \item This work has explored the architecture of INNs to a limited extent. In the field of Computer Vision, there have been numerous studies that investigate various architectures of flow-based INNs \cite{muller2019neural, chen2020residual, stimper2023normflows}. These studies demonstrate the potential for further performance improvements by exploring other INN-based architectural choices.
    

\end{enumerate}

\bibliography{references}
\bibliographystyle{acl_natbib}

\appendix
\clearpage
\onecolumn
\newpage
\appendix
\section{Overview}
\begin{figure*}[ht!]
\begin{center}
    \includegraphics[scale=0.15]{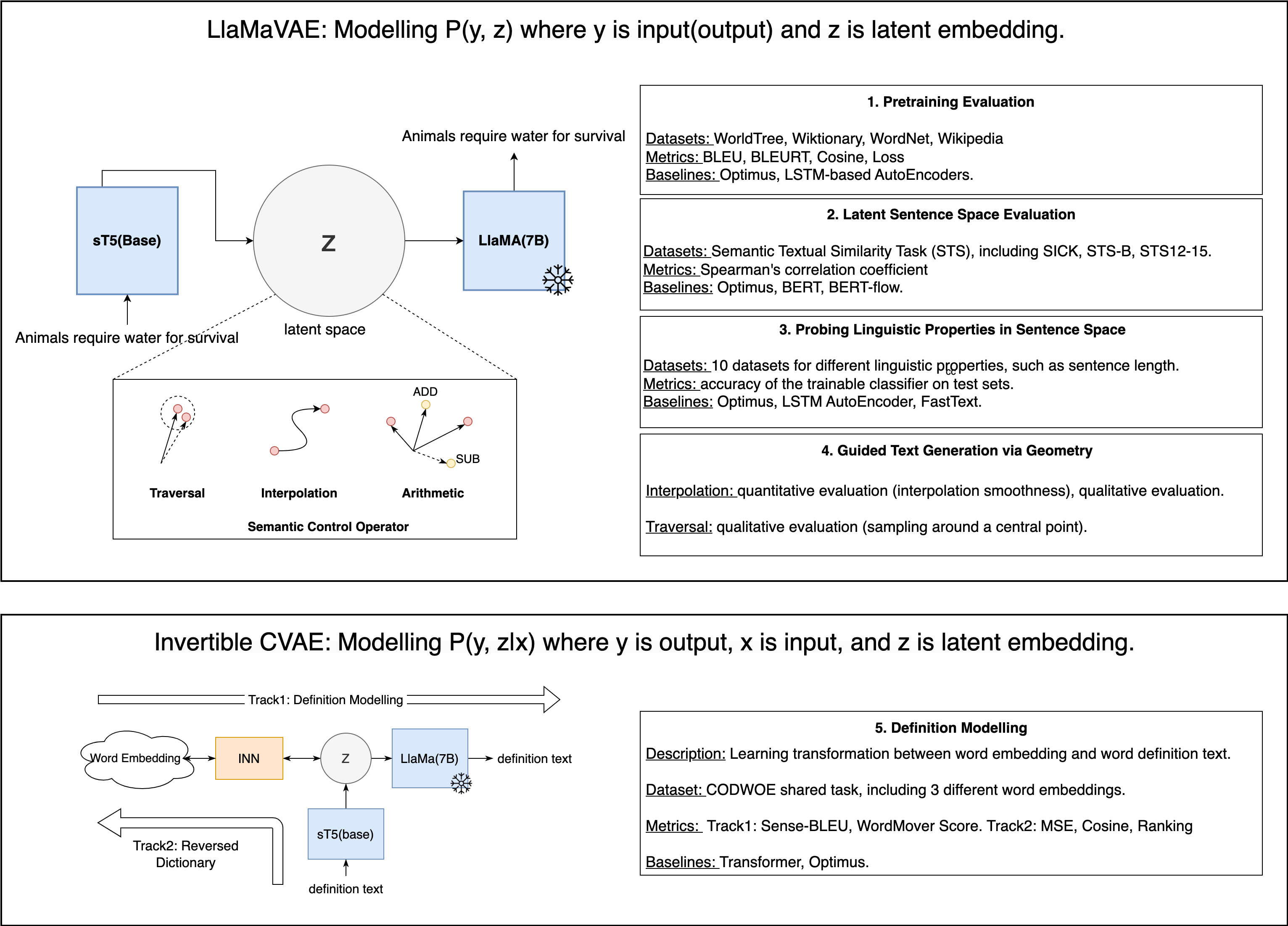}
    \caption{Model architecture and experimental setup.}
    \label{fig:setup}
 \end{center}
\end{figure*}

Figure \ref{fig:setup} displays the model architecture and experimental setup.
\section{VAE implementation} \label{sec:train_details}
The pretrained weights of sT5 (base, mean)\footnote{\url{https://huggingface.co/sentence-transformers/sentence-t5-base}} and LlaMA (7B)\footnote{\url{https://huggingface.co/decapoda-research/llama-7b-hf}} can be acquired from HuggingFace library. The implementation of Optimus is based on their original code \cite{li2020optimus} \footnote{\url{https://github.com/ChunyuanLI/Optimus}}. It is initialised with pretrained Bert and GPT2. The maximal epoch and learning rate are 30 and 5e-04, respectively. The dataset of the definition modelling task can be downloaded via \textit{saf datasets} library, following the command: 

\url{pip install git+https://github.com/neuro-symbolic-ai/saf_datasets.git}

\noindent Then, the dataset can be imported as follows:
\begin{verbatim}
from saf_datasets import CODWOEDataSet
state dataset = CODWOEDataSet()
print(dataset[0].surface)
print(dataset[0].annotations \
["emb_electra"])
\end{verbatim}

\section{INN implementation} \label{sec:inn}
\begin{figure}[ht!]
\begin{center}
    \includegraphics[scale=0.2]{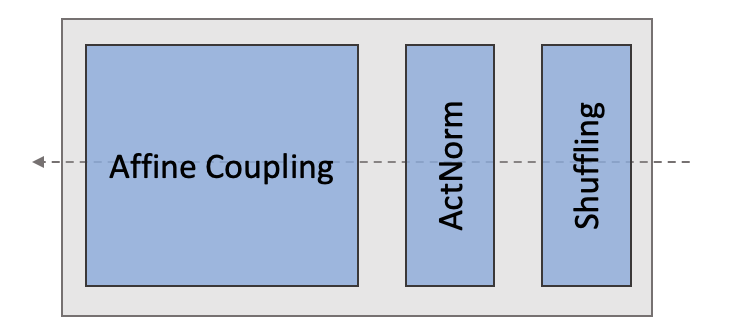}
    \caption{INN one single block.}
    \label{fig:inn_block}
 \end{center}
\end{figure}

INN is implemented via the FrEIA library \cite{freia} \footnote{\url{https://github.com/VLL-HD/FrEIA}}. It consists of 20 invertible blocks. Each of them is built from three layers, including an affine coupling \cite{dinh2016density}, permutation layer, and ActNorm \cite{kingma2018glow}. One block is displayed in Figure \ref{fig:inn_block}. Secondly, we use AdamW \cite{https://doi.org/10.48550/arxiv.1711.05101} to optimize the model where the learning rate is 5e-04 in the experiment. 

\noindent The forward process of the affine coupling layer can be described as follows: 
\begin{equation}
\begin{split}
x_a, x_b = \text{split}(x) \\
\log s, t = m_{\theta}(x_b) \\
s = \exp (\log s) \\
y_a = s \odot x_a + t \\
y_b = x_b \\
y = \text{concat}(y_a, y_b)
\end{split}
\end{equation}
Where $m_{\theta}$ is a two-layer neural network with a dropout = 0.5. $x$ and $y$ are the input and output. The reversed process is:
\begin{equation}
\begin{split}
y_a, y_b = \text{split}(y) \\
\log s, t = m_{\theta}(y_b) \\
s = \exp (\log s) \\
x_a = (y_a - t) / s \\
x_b = y_b \\
y = \text{concat}(x_a, x_b)
\end{split}
\end{equation}
The training process is described in Algorithm \ref{alg:inn_process}.
\begin{algorithm}
    \caption{INN Training Procedure} \label{alg:inn_process}
    \begin{algorithmic}
        \ForAll{batch} \\
          with torch.no\_grad(): \\
            ~~~~ $\mu$, $\Sigma$ = LlaMaVAE(batch) \\
            ~~~~ embed = cat(embed $\times$ 3) \\
            if forward\_train: \\
            ~~~~ z = INN(embed) \\
            ~~~~ loss = 0.5 $\times$ sum((z-$\mu$)$^2$ / $\Sigma$)  \\
            else: \\
            ~~~~ pred\_embed = INN(z, rev=True) \\
            ~~~~ loss = MSE(pred\_embed, embed) \\
        loss.backward()
        \EndFor
    \end{algorithmic}
\end{algorithm}

\begin{figure}[ht!]
\begin{center}
    \includegraphics[scale=0.4]{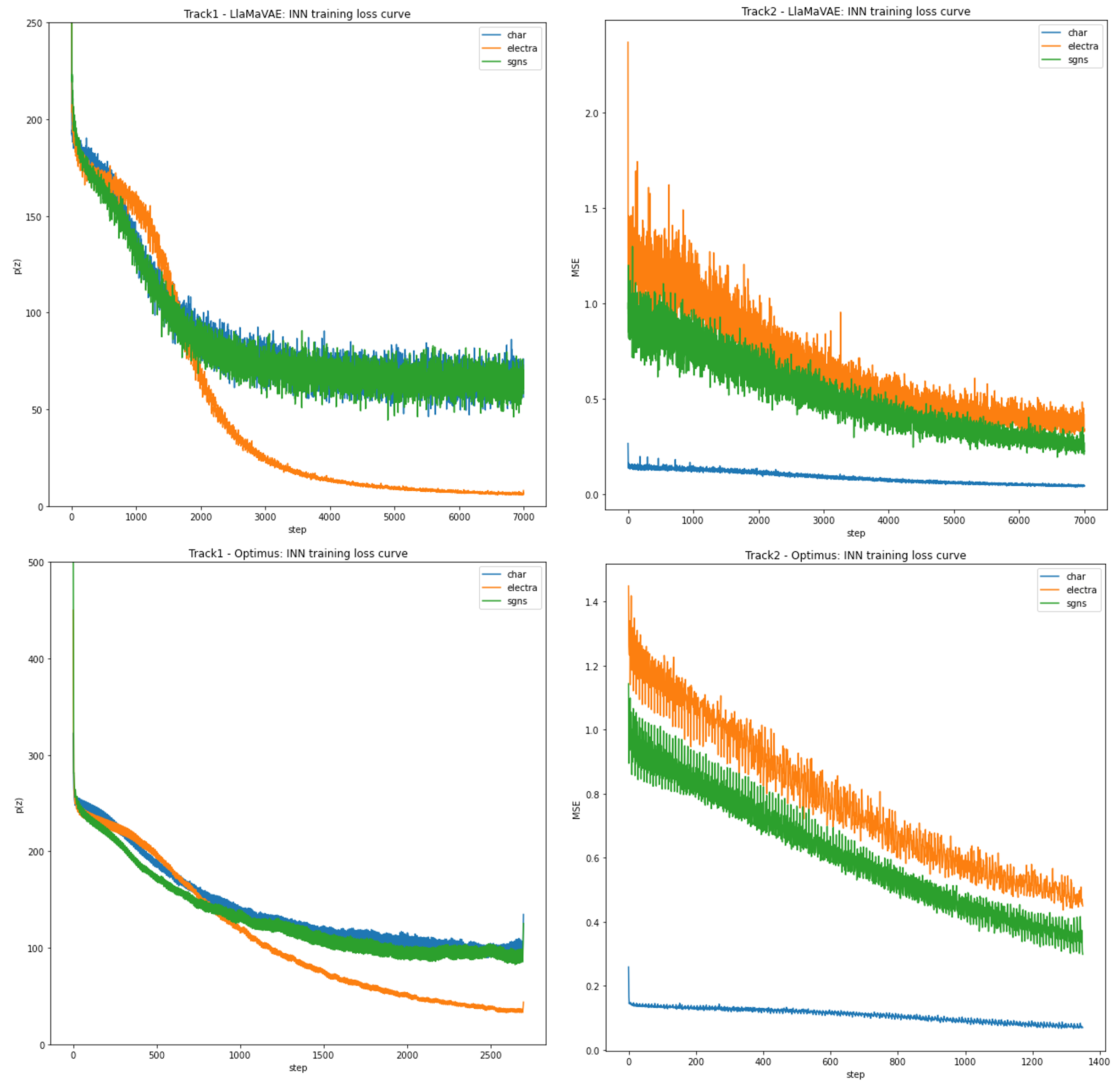}
    \caption{Training loss curve of Definition Modelling task (top: LlaMaVAE, bottom: Optimus, left: track1, right: track2).}
    \label{fig:inn_loss}
 \end{center}
\end{figure}

\noindent The training loss curve of INN in definition modelling is visualized in Figure \ref{fig:inn_loss}.

\section{Interpolation} \label{sec:interpolate_more}

\begin{figure}[ht!]
\begin{center}
    \includegraphics[scale=0.6]{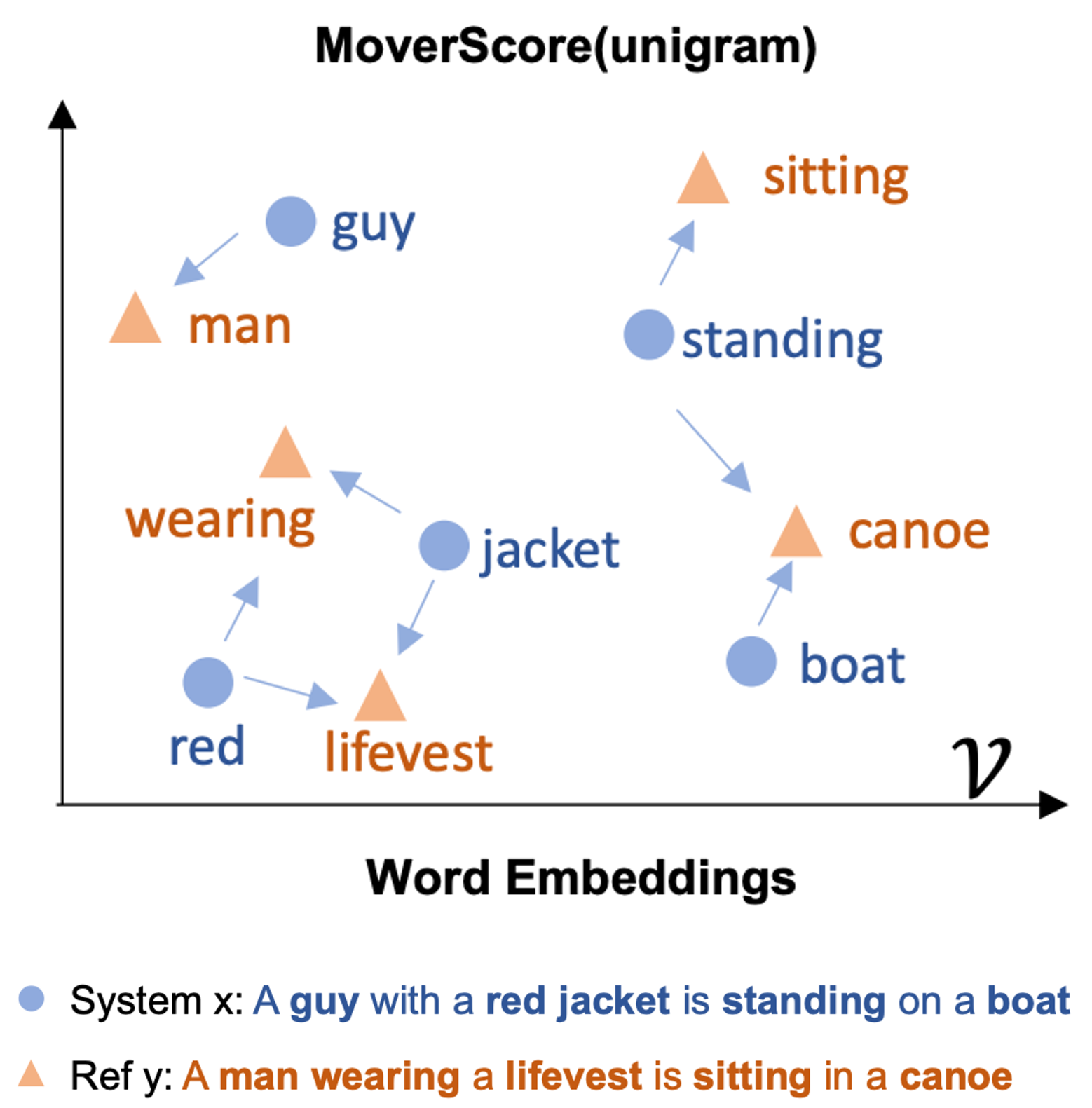}
    \caption{Calculating sentence similarity via word mover's distance \cite{zhao-etal-2019-moverscore}.}
    \label{fig:word_move}
 \end{center}
\end{figure}

\paragraph{Interpolation smoothness} It can be described as the next mathematical equation.
\[
\begin{aligned}
\text{IS} = \mathbb{E}_{(s_0, ..., s_T) \sim P} \frac{\delta(\text{align}(s_0, s_T))}{\sum^T_{t=0} \delta(\text{align}(s_t, s_{t+0.1}))}
\end{aligned}
\]
\noindent Where $s_t$ is the generated sentence at step $t$ in a path, $\delta$ and $\text{align}$ are sentence similarity and alignment functions, respectively. In this experiment, sentence similarity and alignment are performed via Word Mover’s Distance \cite{zhao-etal-2019-moverscore} since it can perform semantic alignment softly. Figure \ref{fig:word_move} displays the process of calculating sentence similarity via word mover's distance.

\paragraph{More interpolation results}
Table \ref{tab:interpolation_optimus} provides the interpolation output for Optimus. Compared with LlaMaVAE in Table \ref{tab:interpolation}, the interpolation of Optimus is not smooth (shown in red colour). Table \ref{tab:interpolation_appendix_1} and \ref{tab:interpolation_appendix} provide more interpolation outputs for both LlaMaVAE and Optimus.
\begin{table}[ht!]
\centering
\begin{tcolorbox}[fontupper=\small, fontlower=\small]
\textcolor{blue}{Source: Mars contains ice} \\
0: Mars contains ice \\
1: \textcolor{red}{rocks are made of hydrogen and helium} \\
2: \textcolor{red}{ice contains water} \\
3: \textcolor{red}{ice cream is made of hydrogen and oxygen} \\
4: Jupiter contains liquid water \\
5: mercury is kind of substance \\
6: food is a kind of simple substance \\
7: food is a kind of substance \\
8: food is a kind of substance \\
9: food is a kind of substance \\
\textcolor{blue}{Target: food is a kind of substance}
\end{tcolorbox}
\caption{Optimus interpolation path where IS is 0.19.}
\label{tab:interpolation_optimus}
\end{table}
\begin{table}[ht!]
\centering
\begin{tcolorbox}[fontupper=\small, fontlower=\small]
\textcolor{blue}{Source: an ice cube is a kind of object} \\
0: an ice cube is a kind of object \\ 
1: ice cube is a kind of object \\
2: ice cube is made of ice \\
3: ice cream is made of water \\
4: ice is made of water \\
5: ice crystals are often made of ice \\
6: ice caps are made of ice \\
7: clouds are formed by water vapor rising \\
8: clouds are formed by water vapor condensing \\
9: clouds are formed by water vapor condensing to form clouds \\

0: an ice cube is a kind of solid \\
1: an ice cube is a kind of object \\
2: an ice cube is a kind of solid formed by an ice cube cooling \\
3: an ice cube is a kind of solid object \\
4: ice is a kind of object \\
5: ice is formed by water vapor rising into colder regions of the atmosphere \\
6: clouds are formed by water vapor rising from oceans \\
7: ice is made of gases trapped in solid ice \\
8: clouds are formed by water vapor evaporating from a source of heat \\
\textcolor{blue}{Target: clouds are formed by water vapor condensing}
\end{tcolorbox}
\caption{Interpolation path (top: LlaMaVAE(IS=0.27), bottom: Optimus(IS=0.21)), only showing unique sentences.}
\label{tab:interpolation_appendix_1}
\end{table}
\begin{table*}[ht!]
\centering
\begin{tcolorbox}[fontupper=\small, fontlower=\small]
\textcolor{blue}{Source: forming sedimentary rock requires burying} \\
0: forming sedimentary rock requires burying \\
1: forming sedimentary rock requires burying \\
2: forming sedimentary rock requires burying sediments \\
3: forming sedimentary rock requires burying the rock \\
4: forming sedimentary rock means sediment is compacted  \\
5: forming a fossil requires the process of burial \\
6: an example of collecting data in an arctic animal ecosystem requires measuring animal habitats \\
7: an example of managing the use of a resource is replacing that resource \\

0: forming sedimentary rock requires compacting and cementing the layers \\
1: creating rocks requires deposition and burial \\
2: forming sedimentary rock requires burying \\
3: forming sedimentary rock requires burying \\
4: forming sedimentary rock requires compacting the materials \\
5: producing something means ( producing ; delivering ) something \\
6: something combining two substances chemically is similar to producing two substances chemically \\
7: an example of managing the use of trees is replacing trees \\
8: an example of managing the use of a resource is replacing that resource \\
9: an example of managing the use of trees is replacing trees \\
10: an example of managing the use of something is using less of that something \\
\textcolor{blue}{Target: an example of managing the use of a resource is replacing that resource} \\

\textcolor{blue}{Source: a cactus wren is a kind of bird}\\
0: a cactus wren is a kind of bird \\
1: a cactus is a kind of plant \\
2: candy is a kind of food \\
3: nutrients are a kind of resource \\
4: sedimentary rock is a kind of rock \\
5: sediment is a kind of material \\

0: a mollusk is a kind of animal \\
1: a cactus wren is a kind of bird \\
2: a cactus stem is a kind of object \\
3: a cactus stem is a kind of plant stem \\
4: boron is a kind of element \\
5: a sedimentary deposit is a kind of deposit \\
6: gravel is a kind of natural material \\
7: sediment is a kind of material \\
\textcolor{blue}{Target: sediment is a kind of material}
\end{tcolorbox}
\caption{Interpolation path (top: LlaMaVAE(IS=0.29, 0.23), bottom: Optimus(IS=0.17, 0.17)).}
\label{tab:interpolation_appendix}
\end{table*}

\end{document}